\documentclass[12pt]{article}
\usepackage[utf8]{inputenc}
\usepackage{apacite}
\usepackage{graphicx}
\usepackage{tikz-dependency}

\title{Response to Liu, Xu, and Liang (2015) and Ferrer-i-Cancho and Gómez-Rodríguez (2015) on Dependency Length Minimization}
\author{Richard Futrell \and Kyle Mahowald \and Edward Gibson}

\begin{document}
\maketitle
\begin{abstract}
  We address recent criticisms (Liu et al., 2015; Ferrer-i-Cancho and Gómez-Rodríguez, 2015) of our work on empirical evidence of dependency length minimization across languages (Futrell et al., 2015).
  First, we acknowledge error in failing to acknowledge Liu (2008)'s previous work on corpora of 20 languages with similar aims.
  A correction will appear in PNAS.
  Nevertheless, we argue that our work provides novel, strong evidence for dependency length minimization as a universal quantitative property of languages, beyond this previous work,
  because it provides baselines which focus on word order preferences.
  Second, we argue that our choices of baselines were appropriate because they control for alternative theories.
\end{abstract}

In recent work, we addressed the question of whether dependency length---the distance between syntactically related words in natural language sentences---is shorter than one would expect under random baselines (Futrell et al., 2015).
This idea has linguistic relevance because if one hypothesizes a universal pressure to minimize dependency length, one can explain a variety of universal properties of languages, including many of the word-order universals noted by Greenberg (1963).
Evidence that language users perfer word orders with shorter dependency length than chance supports this hypothesis, known as the dependency length minimization (DLM) hypothesis.
The DLM hypothesis is theoretically attractive because it is motivated by general human information processing constraints: minimizing dependency length minimizes the online memory load for human sentence parsing and generation.

Two recent articles have raised important criticisms of our work (Liu et al., 2015; Ferrer-i-Cancho \& Gómez-Rodríguez, 2015).

\section{Random Trees and Random Word Orders}

First, Liu et al. (2015) note correctly that we failed to cite a previous large-scale empirical study with similar aims.
In particular, Liu (2008) compares average dependency length in attested sentences of 20 languages to dependency length in random trees.
Not acknowledging this important prior work was an error on our part.
The reason for this omission is that, in all honesty, we did not fully understand this paper and its relationship to ours until conversations with Liu and colleagues after publication.
But these are not good reasons: we acknowledge that we should have made more of an effort to understand and acknowledge prior similar work.
Consequently, we apologize and we urge anyone pursuing research relating to our paper to also study Liu (2008).
This prior work will be acknowledged in a correction to the PNAS article.

\begin{figure}
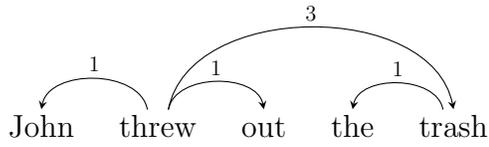

\begin{dependency}[theme=simple]
    \begin{deptext}[column sep=1em]
      John \& threw \& out \& the \& trash \\
    \end{deptext}
    \depedge[label style={font=\normalsize}]{2}{1}{1}
    \depedge[label style={font=\normalsize}]{2}{3}{1}
    \depedge[label style={font=\normalsize}]{2}{5}{3}
    \depedge[label style={font=\normalsize}]{5}{4}{1}
\end{dependency}

Total dependency length = 6 \\

\caption{A possible sentence with its dependency tree and sum dependency length.}
\label{fig:examplesent}
\end{figure}

\begin{figure}
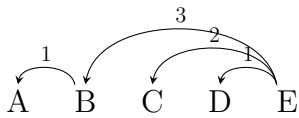

  
\begin{dependency}[theme=simple]
  \begin{deptext}[column sep=1em]
    A \& B \& C \& D \& E \\
  \end{deptext}
  \depedge[label style={font=\normalsize}]{1}{3}{2}
  \depedge[label style={font=\normalsize}]{3}{4}{1}
  \depedge[label style={font=\normalsize}]{3}{5}{2}
  \depedge[label style={font=\normalsize}]{5}{2}{3}
\end{dependency}

Total dependency length = 8 \\

\begin{dependency}[theme=simple]
  \begin{deptext}[column sep=1em]
    A \& B \& C \& D \& E \\
  \end{deptext}
  \depedge[label style={font=\normalsize}]{1}{3}{2}
  \depedge[label style={font=\normalsize}]{1}{5}{4}
  \depedge[label style={font=\normalsize}]{4}{1}{3}
  \depedge[label style={font=\normalsize}]{4}{2}{2}
\end{dependency}

Total dependency length = 11 \\

\begin{dependency}[theme=simple]
  \begin{deptext}[column sep=1em]
    A \& B \& C \& D \& E \\
  \end{deptext}
  \depedge[label style={font=\normalsize}]{2}{1}{1}
  \depedge[label style={font=\normalsize}]{5}{2}{3}
  \depedge[label style={font=\normalsize}]{5}{3}{2}
  \depedge[label style={font=\normalsize}]{5}{4}{1}
\end{dependency}

Total dependency length = 7 \\

\caption{Some random trees based on the sentence in Figure~\ref{fig:examplesent}
  according to the Liu (2008) random tree baseline.}
\label{fig:randomtrees}
\end{figure}

Nevertheless, we believe the difference between the Liu (2008) baselines and ours is non-trivial, such that our work represents new large-scale evidence for the DLM hypothesis.
Liu (2008) uses a ``random tree'' baseline, comparing dependency length in attested dependency trees to dependency length in random ordered trees with the same numbers of nodes.
For example, the dependency length of a sentence with a tree such as in Figure~\ref{fig:examplesent} is compared to the dependency length induced by random ordered trees as in Figure~\ref{fig:randomtrees}.
The baseline trees do not share any syntactic structure with the attested trees they are compared to, beyond their length.
In contrast, Gildea \& Temperley (2010) and Futrell et al. (2015) use ``random word order'' baselines, keeping the syntactic dependency structure of attested sentences constant and investigating random word orders given that syntactic structure, subject to a number of linguistic constraints.
For example, dependency length for a sentence such as in Figure~\ref{fig:examplesent} is compared to dependency length in a sentence with different word order but the same (unordered) dependency tree structure, as in Figure~3.
Attested dependency length is shorter than both the random tree and random word order baselines.

\begin{figure}
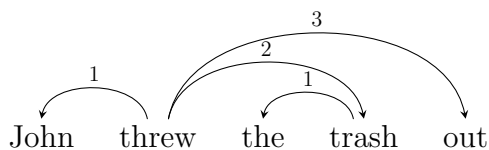

\begin{dependency}[theme=simple]
    \begin{deptext}[column sep=1em]
      John \& threw \& the \& trash \& out \\
    \end{deptext}
    \depedge[label style={font=\normalsize}]{2}{1}{1}
    \depedge[label style={font=\normalsize}]{2}{4}{2}
    \depedge[label style={font=\normalsize}]{4}{3}{1}
    \depedge[label style={font=\normalsize}]{2}{5}{3}
\end{dependency}

Total dependency length = 7 \\

\caption{A random permutation of the sentence in Figure~\ref{fig:examplesent} according to a random word order baseline, specifically the head-fixed projective baseline in Futrell et al. (2015). This baseline permutes sister nodes while maintaining head direction.}
\label{fig:randomorders}
\end{figure}

Our finding that attested dependency length is shorter than random word order baselines shows that, \emph{given} a syntactic structure, language users and language grammars tend to prefer the word order that minimizes dependency length.
This finding supports the DLM hypothesis and provides direct evidence for a specific mechanism (word order preferences) by which dependency length minimization is accomplished.

On the other hand, the finding that attested dependency length is shorter than the random \emph{tree} baselines supports the DLM hypothesis in a more general form and is consistent with many possible mechanisms that shorten dependency length, including non-syntactic mechanisms.
For example, it is consistent with the idea that languages might disprefer structures which inevitably create long dependencies, such as high arity trees.
It is also consistent with the hypothesis that language users prefer sentences with structures that create long dependencies, and might structure discourse to avoid such sentences.
For example, the sentence (1) ``A man who was wearing a hat arrived'' has a long dependency between the subject ``man'' and the verb ``arrived'' because the relative clause ``who was wearing a hat'' intervenes between them.
Language users might prefer to instead say (2) ``A man arrived'', avoiding the relative clause between the subject and the verb, and perhaps mentioning the information about the hat in another sentence later in discourse, or perhaps dropping it altogether.
Though language users are ultimately achieving the same or similar communicative goals in saying sentence (1) and sentence (2), they are doing so by expressing different propositional content in each sentence.
The mechanisms by which dependency length minimization is accomplished in comparison to a random tree baseline are thus highly general: in addition to word order preferences, languages might have tree structure preferences; and language users might strategically choose \emph{what content to express}, in addition to what word order to use, in order to avoid long dependencies.

In summary, comparing to random tree baselines can show DLM as a result of many mechanisms, including the content that people choose to express and/or the word orders they use in sentences.
So the finding that attested dependency length is shorter than this baseline supports an influence of DLM on discourse structure or syntactic structure or both.
Comparing to the random word order baseline, on the other hand, shows specifically that the word orders that people prefer, \emph{given} the content they choose to express, are those that minimize dependency length.
That is, it shows unambiguously that DLM as a pressure affects syntactic structure and word order in particular.
Because our findings are \emph{only} compatible with dependency-length-minimizing preferences in word order, we believe they provide novel, strong evidence for the DLM hypothesis as it pertains to syntax.
Our claim is that, all else being equal, language users prefer linearizations with short dependency length.
Only the comparison to a random word order baseline supports this claim unambiguously.
So we see this work as a complement of Liu (2008) and related work, strengthening the body of evidence for the DLM hypothesis, rather than a repetition.

The difference between random tree baselines and random word order baselines can also explain some discrepancies between our work and previous findings.
For example, we find relatively long dependency lengths for head-final languages such as Japanese and Turkish, whereas Hiranuma (1999) finds that dependency length in Japanese is highly optimized.
Hiranuma (1999)'s finding is specifically that Japanese speakers drop verbal arguments to achieve dependency length minimization, trusting that the language comprehender will be able to infer the missing arguments from discourse context.
Our finding is that, \emph{given} the set of words and the dependency tree that Japanese speakers want to express, they choose orders with longer dependency length than, say, English speakers.
(This finding remains unexplained.)

\section{Projective Baselines}

The second major issue raised in both Liu et al. (2015) and Ferrer-i-Cancho \& Gómez-Rodríguez (2015) is our choice of baselines for comparison.
We use projective linearizations, meaning that when a dependency tree is drawn over a linearized sentence, none of the arcs of the tree cross.
We also use linearizations incorporating other factors that might conceivably influence word order:
a pressure for fixed word order, and a pressure for consistency in head direction.
These three factors---projectivity, head direction consistency, and fixed word order---all have the effect of reducing dependency length, and so it has been argued for the first two that they need not be considered separate factors, but rather the result of DLM.
Ferrer-i-Cancho and Gómez-Rodríguez (2015) argue that our use of these baselines is redundant for this reason.

We believe comparison to these baselines provides stronger evidence for DLM than comparison only to a fully nonprojective baseline,
because it shows that the phenomenon of short dependencies must be explained \emph{even if} independent factors affecting word order are assumed.
Since DLM can explain the phenomena attributed to these other factors, the most parsimonious theory seems to be that DLM is the only factor influencing word order.
But we can only make this argument after showing that the shortness of dependencies persists as a phenomenon even after controlling for these other hypothetical factors.
For example, suppose we had found that attested dependency length was \emph{not} shorter than the projective random baselines\footnote{Which would not have been surprising given previous work: Gildea \& Temperley (2010) found much weaker minimization in German than in English.}.
One would be left with the question of why, if DLM is the main factor influencing language structure, German speakers pass up opportunities to minimize dependency length.
Then one could argue that DLM is not a good explanation for projectivity, since word orders are not minimized for dependency length beyond what is needed to establish projectivity, which itself might have independent motivations (such as enabling polynomial-time parsing).
Since we found that dependency length \emph{is} shorter than this baseline in many languages, this line of argumentation is no longer available.

\begin{figure}
  \centerline{\includegraphics[scale=.55]{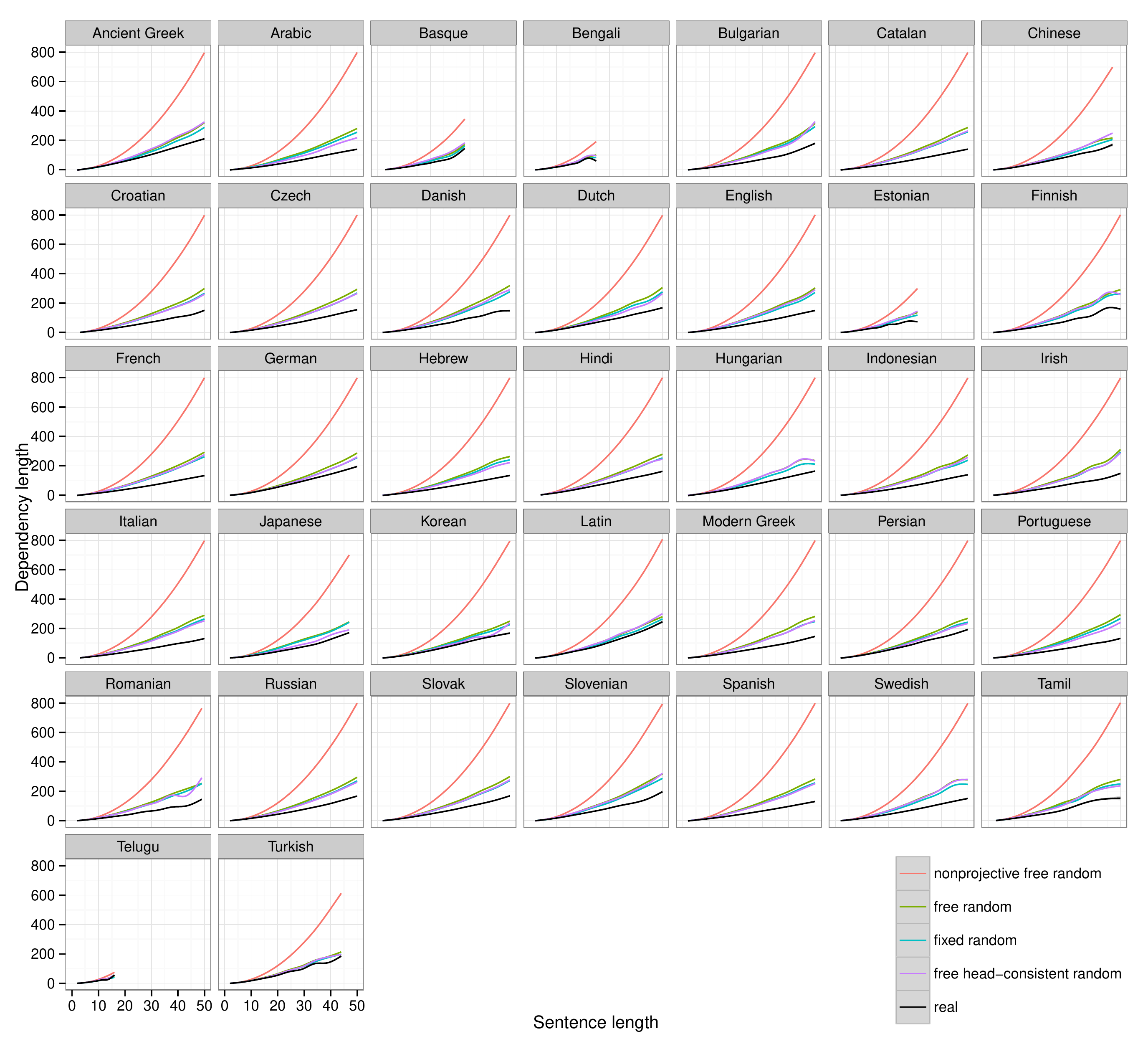}}
  \caption{Dependency length as a function of sentence length, for real sentences (black), the free nonprojective baseline (red), and several baselines from the paper. All data except for the free nonprojective baseline were present in the original paper.}
  \label{fig:nonproj}
\end{figure}

For the sake of completeness, we provide a comparison of attested dependency lengths with dependency lengths in random nonprojective linearizations in Figure~\ref{fig:nonproj}.
For this baseline, the dependency tree is linearized by shuffling nodes at random.
The baselines from Futrell et al. (2015) are also shown.
The figure shows that dependency length is much shorter than the nonprojective baseline, and that the projective baselines are much more conservative than the nonprojective baseline.
We felt that including the nonprojective baselines in the original paper would be redundant, since Ferrer-i-Cancho (2006) showed that projective trees on average have shorter dependency length than nonprojective trees, and Kuhlmann (2013) (among others) showed that natural language dependency trees are overwhelmingly projective.

We also want to stress that, contra Ferrer-i-Cancho \& Gómez-Rodríguez (2015), controlling for these possible alternative factors affecting word order does not imply that we are accepting traditional nativist or Universal Grammar-based hypotheses.
These factors have possible functional explanations, just as DLM does.
Fixed word order can be motivated by efficient communication of relation types;
consistent head direcion can be motivated by compression of grammars;
and projectivity can be motivated by the time complexity of parsing, where parsing to projective trees is cubic-time but parsing to fully nonprojective trees is NP-hard.
In general, we aimed to include the most conservative reasonable baselines.

\section{Other Issues}

Liu et al. (2015) also raise a number of more specific criticisms.
They claim that the uniformity of genres of the text in our corpora could be a confounding factor.
The criticism is valid: It is true that our corpora were primarily (but not entirely) written text from newspapers and novels.
Nevertheless, we would find it surprising if DLM universally influenced novels and newspapers but not language use in general.
We welcome any work which controls for this possible issue.

Finally, Liu et al. (2015) also note that in our original paper we state that head-final languages appear to have longer dependencies than more head-initial or head-medial languages, but we do not provide statistical tests of this claim.
We intended this remark not as a main claim of the paper, but as a conjecture intended to draw attention to the wide variation between languages in their dependency length, and possible typological implications of that variation.
Working out the correct statistical methodology and gathering the right data to make this a strong empirical claim would require another whole paper.
The question of variation in dependency length has also been a major focus of Liu's research.
We feel that explaining this variation is the most interesting direction for future dependency length research,
and we hope to join our present critics in future investigations of this phenomenon.

\end{document}